\documentclass[11pt]{article}

\usepackage{amsmath, amssymb, amsthm}
\usepackage[margin=1in]{geometry}
\usepackage{graphicx}
\usepackage{booktabs}
\usepackage[authoryear]{natbib}
\usepackage{xcolor}
\usepackage{listings}
\usepackage{caption}
\usepackage{xurl}
\usepackage{hyperref}
\hypersetup{
    colorlinks=true,
    linkcolor=blue,
    citecolor=blue,
    urlcolor=blue
}

\providecommand{\doi}[1]{\href{https://doi.org/#1}{doi: #1}}

\newtheorem{proposition}{Proposition}

\theoremstyle{definition}

\title{A PyTorch Framework for Scalable Non-Crossing Quantile Regression}

\author{
    Kaihua Chang \\
    Arizona Department of Education \\
    {\small \href{mailto:Kaihua.Chang@azed.gov}{\texttt{Kaihua.Chang@azed.gov}}}
}

\date{December 2025}

\begin{document}

\maketitle

\begin{abstract}
Quantile regression is fundamental to distributional modeling, yet independent estimation of multiple quantiles frequently produces crossing---where estimated quantile functions violate monotonicity, implying impossible negative probability densities. While Constrained Joint Quantile Regression (CJQR) elegantly enforces non-crossing by construction, existing formulations via Linear Programming exhibit $O((qn)^3)$ complexity, rendering them impractical for large-scale applications. We present the first scalable solution using PyTorch automatic differentiation: \textbf{CJQR-ALM}, combining the \textbf{Augmented Lagrangian Method} with \textbf{differentiable pinball loss} and \textbf{L-BFGS} optimization. Our approach reduces computational complexity to $O(n)$, achieving near-zero crossing rates on datasets exceeding 70,000 observations within minutes. The differentiable formulation naturally extends to neural network architectures for non-linear conditional quantile estimation. Application to Student Growth Percentile calculations demonstrates practical utility for educational assessment, while simulation studies show negligible accuracy cost (RMSE increase $\approx 2.4$ points) relative to unconstrained estimation---a favorable trade-off for applications requiring valid probability statements across finance, healthcare, and engineering.
\end{abstract}

\section{Introduction}

Quantile regression \citep{koenker1978regression, koenker2005quantile} has become essential for modeling conditional distributions across economics, finance, healthcare, and education. Unlike mean regression, quantile regression characterizes the entire conditional distribution by estimating multiple quantile functions $Q_{\tau}(x)$ for $\tau \in (0,1)$.

A fundamental challenge arises when quantiles are estimated independently: nothing guarantees that $\hat{Q}_{\tau_1}(x) \leq \hat{Q}_{\tau_2}(x)$ for $\tau_1 < \tau_2$. This \emph{quantile crossing} phenomenon \citep{koenker2005quantile, dette2008estimating, chernozhukov2010quantile} produces estimated distributions with locally negative probability densities---a mathematical impossibility that undermines the interpretability of the entire distributional estimate.

\subsection{The Crossing Problem}

Consider estimating $q = 100$ conditional quantiles for a dataset with $n = 70{,}000$ observations---a common scenario in educational assessment, insurance modeling, or financial risk analysis. When each quantile is fitted independently, crossing can affect a substantial proportion of observations. For heteroscedastic data with complex conditional distributions, crossing rates can exceed 80\% of observations under strict definitions (any adjacent quantile pair violation).

The standard solution is \emph{isotonization}: post-hoc sorting of quantile estimates at each covariate point \citep{chernozhukov2010quantile}. While computationally efficient and asymptotically justified, isotonization raises concerns for finite-sample applications:
\begin{enumerate}
    \item The isotonized estimates no longer minimize the original quantile loss
    \item When a large majority of cases require correction, isotonization performs reconstruction rather than adjustment
    \item The underlying model produced structurally invalid probability statements that required repair
\end{enumerate}

\subsection{Constrained Joint Quantile Regression}

\citet{bondell2010noncrossing} proposed an elegant alternative: estimate all quantiles jointly with explicit non-crossing constraints:
\begin{align}
\min_{\{\beta(\tau_j)\}} &\sum_{j=1}^q \sum_{i=1}^n \rho_{\tau_j}(Y_i - \Phi(X_i)^\top \beta(\tau_j)) \label{eq:cjqr_obj} \\
\text{subject to} \quad &\Phi(X_i)^\top \beta(\tau_j) \leq \Phi(X_i)^\top \beta(\tau_{j+1}) \quad \forall i, j \label{eq:cjqr_constraint}
\end{align}
where $\rho_\tau(u) = u(\tau - \mathbb{I}_{u<0})$ is the pinball loss and $\Phi(x)$ contains basis functions (e.g., B-splines).

This formulation guarantees non-crossing by construction. However, the resulting Linear Programming (LP) problem has $O(qn)$ variables and $O(qn)$ constraints, leading to $O((qn)^3)$ complexity for interior point methods---prohibitive for datasets with tens of thousands of observations.

\subsection{Motivation}

This work was motivated by Student Growth Percentiles (SGP), a methodology introduced by \citet{betebenner2009norm} that has become predominant for quantifying academic progress in U.S. educational accountability systems. SGP estimates 100 conditional quantiles of current achievement given prior achievement, enabling normative comparisons of student growth trajectories. The standard implementation addresses crossing through isotonization---post-hoc sorting that enforces monotonicity.

During operational implementation, we observed that crossing can affect a substantial proportion of students in certain cohorts, particularly those with heteroscedastic conditional distributions or ceiling effects in prior scores. While isotonization successfully repairs these estimates, we sought an alternative that produces valid estimates directly.

Additionally, the migration of assessment pipelines from R to Python revealed ecosystem limitations. The R \texttt{quantreg} package relies on highly optimized Fortran implementations of the Barrodale-Roberts simplex algorithm \citep{koenker1987computing} and the Frisch-Newton interior point method \citep{portnoy1997gaussian}. Python's standard alternative, the Iteratively Reweighted Least Squares (IRLS) algorithm in statsmodels, exhibited convergence instability and prohibitive computational overhead when applied to large-scale operational data, frequently failing to reach tolerance thresholds compared to the specialized Frisch-Newton interior point methods used in R.

Rather than re-engineering legacy LP solvers, we utilized the comparative advantage of the modern Python ecosystem: \textbf{automatic differentiation}. By reformulating the quantile objective within PyTorch, we bypass the complexity of approximating the Frisch-Newton method entirely. This shift from discrete LP solvers to continuous gradient-based optimization (L-BFGS) not only restores computational parity but enables the scalable solution of the \emph{constrained joint estimation} problem---transforming a software migration task into a methodological contribution.

\subsection{Contributions}

This paper bridges the gap between the theoretical elegance of constrained joint estimation and the computational demands of large-scale applications:

\begin{enumerate}
    \item \textbf{First Scalable Non-Crossing Solution via Automatic Differentiation:} We present \textbf{CJQR-ALM}, the first implementation combining the Augmented Lagrangian Method with differentiable pinball loss and L-BFGS optimization for non-crossing quantile regression. Unlike prior LP-based approaches \citep{bondell2010noncrossing} that exhibit $O((qn)^3)$ complexity, our PyTorch-based formulation achieves $O(n)$ complexity, enabling practical application to datasets with tens of thousands of observations.

    \item \textbf{PyTorch Framework for Modern ML Pipelines:} Our implementation leverages automatic differentiation, enabling seamless integration with deep learning workflows. The differentiable formulation naturally extends to neural network architectures for non-linear conditional quantile estimation, opening pathways to deep distributional learning with guaranteed monotonicity.

    \item \textbf{Accuracy-Validity Trade-off Characterization:} Through simulation, we quantify the cost of enforcing structural validity. CJQR-ALM achieves near-zero crossing at a negligible accuracy penalty (RMSE increase $\approx 2.4$ points)---an order of magnitude smaller than typical measurement error.
    
    \item \textbf{Cross-Domain Applicability:} While motivated by educational assessment (SGP), the framework applies broadly to any domain requiring valid distributional estimates:
    \begin{itemize}
        \item \textbf{Finance:} Value-at-Risk (VaR) and Conditional VaR (CVaR) estimation with guaranteed monotonicity
        \item \textbf{Healthcare:} Medical outcome prediction intervals that respect probability axioms
        \item \textbf{Insurance:} Risk quantification across the full loss distribution
        \item \textbf{Engineering:} Reliability analysis with coherent failure probability estimates
    \end{itemize}
\end{enumerate}

\section{Mathematical Framework}

\subsection{Quantile Regression Fundamentals}

The $\tau$-th conditional quantile of $Y$ given $X = x$ satisfies:
\begin{equation}
P(Y \leq Q_\tau(x) | X = x) = \tau
\end{equation}

Under a linear specification with basis expansion \citep{deboor1978splines}, $Q_\tau(x) = \Phi(x)^\top \beta(\tau)$. The standard estimator minimizes the pinball loss \citep{koenker1978regression}:
\begin{equation}
\hat{\beta}(\tau) = \arg\min_{\beta} \sum_{i=1}^n \rho_\tau(Y_i - \Phi(X_i)^\top \beta)
\end{equation}
where $\rho_\tau(u) = u(\tau - \mathbb{I}_{u<0})$ is convex but non-differentiable at the origin.

\subsection{The Crossing Problem}

When quantiles are estimated independently, crossing occurs when $\hat{Q}_{\tau_1}(x) > \hat{Q}_{\tau_2}(x)$ for some $x$ despite $\tau_1 < \tau_2$. This implies the estimated conditional CDF is non-monotonic, producing locally negative probability densities.

For applications requiring valid distributional interpretations---such as growth percentiles, value-at-risk, or prediction intervals---crossing represents a fundamental validity failure, not merely a numerical inconvenience.

\subsection{Problem Formulation}

We formulate the Constrained Joint Quantile Regression problem as:
\begin{align}
\min_{\{\beta(\tau_j)\}_{j=1}^q} \quad &\sum_{j=1}^q \sum_{i=1}^n \rho_{\tau_j}(Y_i - \Phi(X_i)^\top \beta(\tau_j)) \\
\text{subject to} \quad &\Phi(X_i)^\top \beta(\tau_j) + \epsilon \leq \Phi(X_i)^\top \beta(\tau_{j+1}) \quad \forall i \in [n], j \in [q-1]
\end{align}
where $\epsilon > 0$ is a small margin ensuring strict separation (typically $\epsilon = 10^{-4}$).

This formulation has:
\begin{itemize}
    \item \textbf{Variables:} $q \cdot p$ parameters, where $p$ is the number of basis functions
    \item \textbf{Constraints:} $n \cdot (q-1)$ non-crossing constraints
\end{itemize}

For typical applications ($n = 70{,}000$, $q = 100$, $p = 10$), this yields 1,000 parameters and nearly 7 million constraints.

\section{Methodology: Augmented Lagrangian Method}

We present a scalable approach using the Augmented Lagrangian Method (ALM) with automatic differentiation. To our knowledge, this represents the first application of ALM to the CJQR problem; prior work has relied on linear programming \citep{bondell2010noncrossing} or post-hoc rearrangement \citep{chernozhukov2010quantile}.

\subsection{Differentiable Pinball Loss}

The standard pinball loss is non-differentiable at $u=0$, precluding quasi-Newton methods like L-BFGS that approximate the inverse Hessian. We employ a \textbf{Huberized Pinball Loss}---a smooth approximation that replaces the kink with a quadratic transition:
\begin{equation}
L_{\tau,\delta}(u) = 
\begin{cases} 
w_\tau \cdot \frac{u^2}{2\delta} & \text{if } |u| \leq \delta \\
w_\tau \cdot (|u| - \frac{\delta}{2}) & \text{otherwise}
\end{cases}
\end{equation}
where $w_\tau = \tau$ if $u \geq 0$ and $w_\tau = 1-\tau$ otherwise.

The smoothing parameter $\delta > 0$ controls the trade-off between smoothness and fidelity to the original objective. We use $\delta = 0.05$ in our implementation, which provides sufficient smoothness for L-BFGS while asymptotically approaching the true quantile objective.

\textbf{Distinction from Standard Huber Loss:} The classical Huber loss \citep{huber1964robust} is symmetric and designed for robust mean estimation. Our Huberized Pinball Loss preserves the asymmetric weighting ($\tau$ vs. $1-\tau$) essential to quantile regression, applying smoothing only to address non-differentiability.

\subsection{Optimizer Selection: L-BFGS}

We selected the Limited-memory BFGS (L-BFGS) algorithm \citep{liu1989lbfgs, nocedal2006} over stochastic methods (SGD, Adam) for two reasons:

\textbf{Rationale 1: Convergence Properties.} For datasets that fit in memory ($N < 10^6$), L-BFGS achieves \emph{superlinear convergence} by approximating the inverse Hessian, requiring significantly fewer iterations than first-order methods with \emph{linear convergence}.

\textbf{Rationale 2: Constraint Stability.} The Augmented Lagrangian Method requires precise gradient computations to update dual variables. Stochastic noise from mini-batch sampling destabilizes these updates, causing oscillation around constraint boundaries rather than precise satisfaction.

\subsection{Augmented Lagrangian Formulation}

Define the constraint violation:
\begin{equation}
g_{ij}(\theta) = Q_{\tau_j}(x_i) - Q_{\tau_{j+1}}(x_i) + \epsilon
\end{equation}

The augmented Lagrangian \citep{hestenes1969multiplier, powell1969method} is:
\begin{equation}
\mathcal{L}(\theta, \mu, \rho) = \sum_{\tau,i} L_{\tau,\delta}(y_i - Q_\tau(x_i)) + \sum_{i,j} \left[ \mu_{ij}[g_{ij}]_+ + \frac{\rho}{2}[g_{ij}]_+^2 \right]
\end{equation}
where $[x]_+ = \max(0, x)$, $\mu_{ij} \geq 0$ are Lagrange multipliers, and $\rho > 0$ is the penalty parameter.

\begin{proposition}
With linear basis $Q_\tau(x)=\Phi(x)^\top\beta(\tau)$, the Huberized pinball loss is convex, and the non-crossing constraints form a linear polytope. The joint problem is convex, and ALM converges to a global optimum under standard regularity conditions \citep{nocedal2006}.
\end{proposition}

\subsection{Two-Phase Optimization Algorithm}

We employ a two-phase approach for robust convergence:

\textbf{Phase 1 (Warm-Up):} Optimize the unconstrained Huberized pinball loss for 100 iterations to establish good initialization. This allows the model to learn basic quantile structure before constraints are enforced.

\textbf{Phase 2 (Constrained ALM):}

\begin{enumerate}
    \item \textbf{Primal Update:} Minimize $\mathcal{L}(\theta, \mu, \rho)$ with respect to $\theta$ using L-BFGS
    \item \textbf{Dual Update:} Update multipliers: $\mu_{ij} \leftarrow \mu_{ij} + \rho [g_{ij}]_+$
    \item \textbf{Penalty Update:} If violation does not decrease sufficiently ($V_{k+1} > \gamma \cdot V_k$ with $\gamma = 0.9$), increase penalty: $\rho \leftarrow \alpha \rho$ with $\alpha = 4.0$
    \item \textbf{Convergence:} Terminate when $\max_{i,j} [g_{ij}]_+ < 10^{-6}$
\end{enumerate}

The adaptive penalty mechanism ensures $\rho$ grows only when necessary, avoiding numerical ill-conditioning while driving violations toward zero.

\subsection{Computational Complexity Analysis}

The reduction from cubic to linear complexity is central to scalability:

\textbf{Standard LP Approach:} Interior point methods for the equivalent LP solve systems where both variables and constraints scale with $n$, yielding $O((qn)^3)$ worst-case complexity.

\textbf{Our ALM Approach:} By operating in parameter space ($V = q \cdot p$) rather than primal-dual space, complexity is dominated by:
\begin{enumerate}
    \item \textbf{Lagrangian Evaluation:} $O(n \cdot q)$ to compute loss and constraint violations
    \item \textbf{L-BFGS Update:} $O(V)$ operations per iteration, where $V = q \cdot p$ is constant relative to $n$
\end{enumerate}

Since $q$ (number of quantiles) and $p$ (basis functions) are fixed constants, per-iteration complexity is $\mathbf{O(n)}$. Typical convergence requires 10--70 outer iterations.

\begin{table}[h]
\centering
\caption{Computational Complexity Comparison}
\label{tab:complexity}
\begin{tabular}{@{}lc@{}}
\toprule
\textbf{Method} & \textbf{Complexity} \\
\midrule
LP Interior Point \citep{bondell2010noncrossing} & $O((qn)^3)$ \\
Post-hoc Isotonization \citep{chernozhukov2010quantile} & $O(nq \log q)$ \\
\textbf{CJQR-ALM (Proposed)} & $O(n)$ \\
\bottomrule
\end{tabular}
\end{table}

\section{Experiments}

We evaluate CJQR-ALM through (1) a demonstration on operational data with known crossing, (2) accuracy benchmarking via simulation, and (3) application to operational Student Growth Percentile (SGP) calculations.

\subsection{Demonstration: Crossing Elimination}

Figure~\ref{fig:crossing_comparison} demonstrates the crossing problem on operational SGP data. The top panel shows that independent quantile regression produces crossing at high prior scores---quantile curves overlap and invert, implying negative probability density in those regions. The bottom panel shows CJQR-ALM estimates, which maintain strict monotonicity ($Q_{\tau_j}(x) < Q_{\tau_{j+1}}(x)$ for all $\tau_j < \tau_{j+1}$) throughout the covariate space.

The ALM optimization proceeds through adaptive penalty updates: constraint violation decreases exponentially toward the tolerance threshold ($10^{-6}$), Lagrange multipliers accumulate at binding constraints (identifying crossing points), and the penalty parameter $\rho$ increases only when violation reduction stagnates.

\subsection{Accuracy Trade-off: Simulation Study}

To benchmark accuracy against ground truth, we conducted a simulation study with known data-generating process.

\textbf{Design:} $N=3{,}000$ observations, $q = 100$ quantiles, 20 replications. The DGP mimics heteroscedastic assessment data:
\begin{equation}
Y = \mu(X) + \sigma(X) \cdot \varepsilon, \quad \varepsilon \sim \mathcal{N}(0,1)
\end{equation}
where $\mu(X) = 2400 + 15X + 0.5X^2$ and $\sigma(X) = 10 + 0.5X$, with $X \sim \text{Uniform}(0, 10)$.

True conditional quantiles $Q_\tau(x) = \mu(x) + \sigma(x) \cdot \Phi^{-1}(\tau)$ are computed analytically, enabling exact RMSE calculation.

\textbf{Results:} Table~\ref{tab:simulation} summarizes the RMSE comparison across 20 replications.

\begin{table}[h]
\centering
\caption{Simulation Study: RMSE Comparison (20 Replications)}
\label{tab:simulation}
\begin{tabular}{@{}lcc@{}}
\toprule
\textbf{Method} & \textbf{Mean RMSE} & \textbf{Range} \\
\midrule
Independent QR + Isotonization & 1.12 & [0.81, 1.40] \\
CJQR-ALM (Proposed) & 3.56 & [3.23, 3.92] \\
\midrule
\textbf{Difference (Validity Tax)} & \textbf{2.44} & --- \\
\bottomrule
\end{tabular}
\end{table}

\textbf{Interpretation:} CJQR-ALM incurs a 2.44-point RMSE penalty---the \textbf{``validity tax''} for ensuring structural validity. This difference is an order of magnitude smaller than typical measurement error (SEM $\approx$ 10--15 points for standardized assessments). RMSE rewards flexibility regardless of structural validity; an estimator with lower RMSE but substantial crossing produces mathematically impossible distributions. The validity tax is negligible for applications where coherent probability statements are required.

\subsection{Application: Student Growth Percentiles}

We applied CJQR-ALM to operational Student Growth Percentile (SGP) calculations on cohorts exceeding 70,000 students. SGP \citep{betebenner2009norm} estimates 100 conditional quantiles of current achievement given prior achievement, making it an ideal testbed for non-crossing methods.

\textbf{Key findings from application:}
\begin{itemize}
    \item CJQR-ALM achieves near-zero crossing on cohorts where independent estimation exhibits substantial crossing
    \item High agreement between CJQR-ALM and isotonized SGP confirms both methods measure the same construct
    \item Runtime for 70,000+ students: 5--10 minutes on standard server hardware
\end{itemize}

\subsection{Convergence Behavior}

\begin{figure}[htbp]
    \centering
    \includegraphics[width=0.9\linewidth]{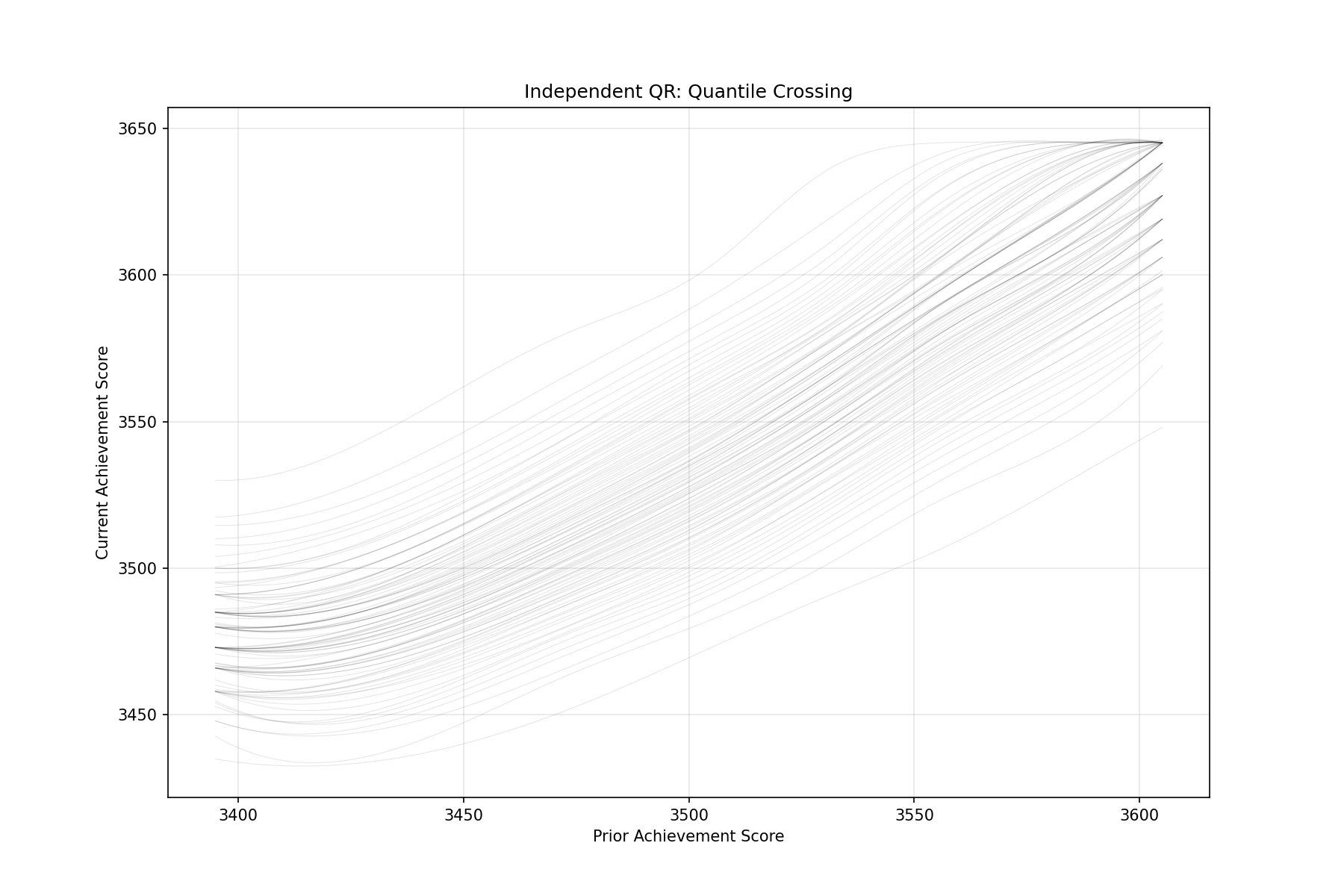}
    \par\vspace{1em}
    \includegraphics[width=0.9\linewidth]{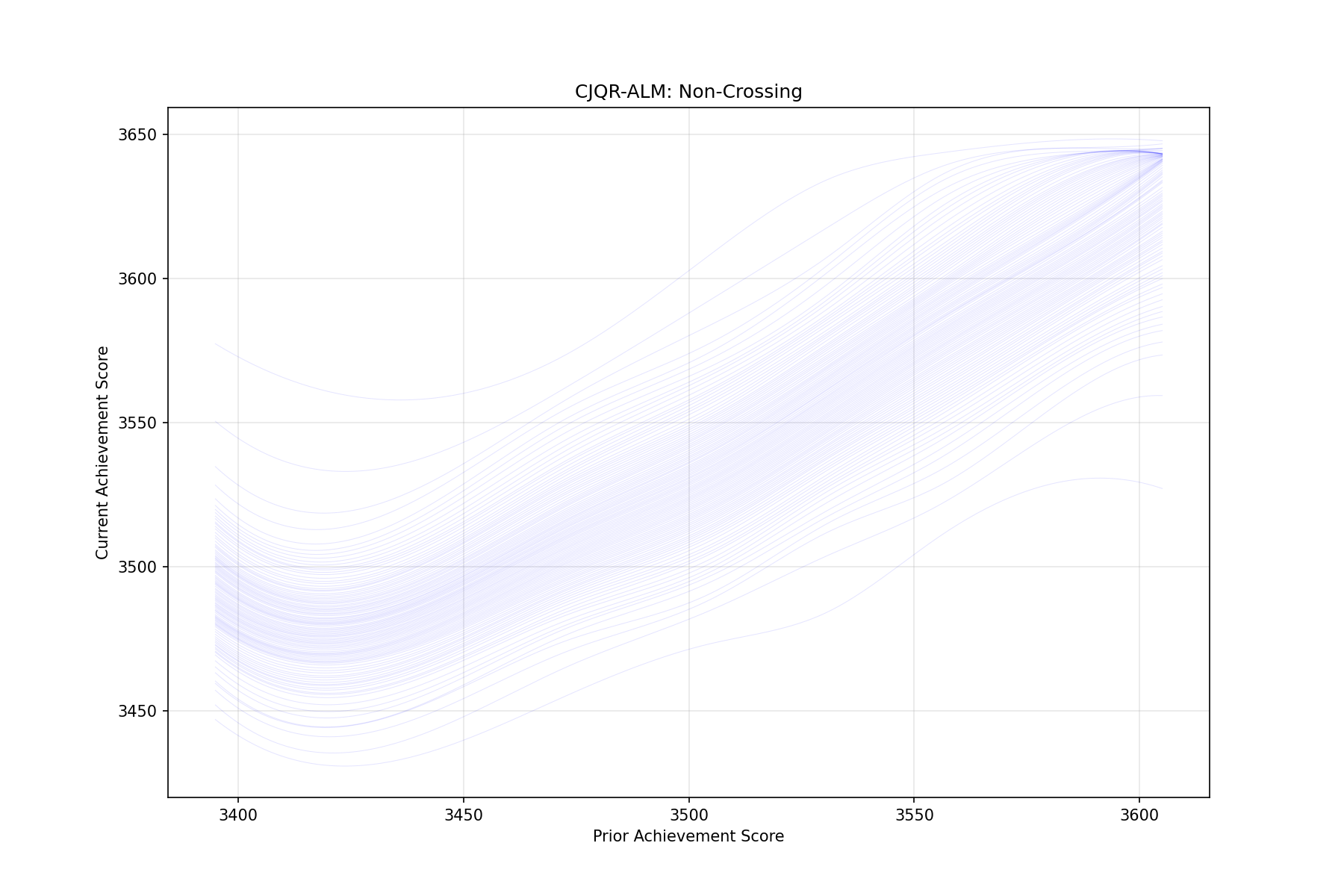}
    \caption{Conditional quantile estimates for Student Growth Percentiles. \textbf{Top:} Independent QR exhibits crossing at high prior scores due to ceiling effects. \textbf{Bottom:} CJQR-ALM enforces strict monotonicity throughout the covariate space.}
    \label{fig:crossing_comparison}
\end{figure}

Figure~\ref{fig:crossing_comparison} illustrates the crossing problem and its resolution on operational SGP data. The top panel shows independent quantile regression estimates, where quantile curves exhibit visible crossing---particularly at high prior scores where ceiling effects concentrate students. The bottom panel shows CJQR-ALM estimates, which maintain strict monotonicity throughout the covariate space.

\section{Discussion}

\subsection{When to Use CJQR-ALM}

CJQR-ALM is appropriate when:
\begin{itemize}
    \item \textbf{Valid probability statements are required:} Applications where each quantile must correspond to a coherent probability (e.g., growth percentiles, prediction intervals, risk measures)
    \item \textbf{Post-hoc correction is undesirable:} Settings where model outputs should be directly interpretable without additional processing
    \item \textbf{Crossing rates are substantial:} When independent estimation produces substantial crossing, the underlying model has failed to capture conditional geometry
\end{itemize}

For applications prioritizing point estimate accuracy over distributional validity, independent estimation with isotonization remains appropriate.

\subsection{Extension to Neural Networks}

The differentiable formulation of CJQR-ALM naturally extends to neural network architectures. By replacing the linear model $Q_\tau(x) = \Phi(x)^\top \beta(\tau)$ with a neural network $Q_\tau(x) = f_\theta(x, \tau)$, the same ALM framework enforces non-crossing constraints on non-linear conditional quantile functions. This enables \textbf{deep distributional learning with guaranteed monotonicity}---a capability not available in existing deep quantile regression implementations that rely on post-hoc sorting or architectural tricks.

\subsection{Relationship to Isotonization}

Isotonization \citep{chernozhukov2010quantile} and CJQR-ALM address the same problem through different mechanisms:

\begin{itemize}
    \item \textbf{Isotonization:} Estimate independently, then sort. Fast and asymptotically justified, but the pre-sorted estimates were structurally invalid.
    \item \textbf{CJQR-ALM:} Estimate jointly with constraints. Slightly higher computational cost, but estimates are valid by construction.
\end{itemize}

In our SGP application, we observed high agreement between methods, confirming both measure the same underlying construct. The choice reflects a philosophical preference: isotonization prioritizes local fit, while CJQR prioritizes global coherence.

\subsection{Limitations}

\textbf{Accuracy cost:} CJQR-ALM pays a small RMSE penalty ($\approx 2.4$ points in our simulations) relative to unconstrained estimation. This cost is negligible for most applications but may matter for research requiring maximal point estimation precision.

\textbf{Convexity requirement:} Our convergence guarantees require convex loss and linear constraints. Extensions to non-convex settings (e.g., deep neural network quantile regression) would require additional algorithmic development, though preliminary experiments suggest the ALM framework remains effective.

\textbf{Coverage properties:} Enforcing strict monotonicity may slightly shift empirical coverage from nominal values at extreme quantiles. Future work could characterize finite-sample coverage.

\section{Related Work}

\textbf{Non-crossing quantile regression:} \citet{bondell2010noncrossing} introduced the constrained joint formulation solved via LP. \citet{dette2008estimating} proposed kernel-based approaches with monotonicity constraints. Our contribution is the first scalable ALM implementation using automatic differentiation, enabling application to large datasets within modern ML frameworks.

\textbf{Post-hoc methods:} \citet{chernozhukov2010quantile} established the theoretical foundations for rearrangement (isotonization), proving asymptotic super-efficiency. We view isotonization as complementary---appropriate when computational cost is paramount; CJQR is preferable when structural validity is paramount.

\textbf{Deep quantile regression:} Recent work on non-crossing neural network quantile regression \citep{zhou2020noncrossing} uses architectural constraints (cumulative sum layers) to enforce monotonicity. Our ALM approach offers an alternative that preserves standard network architectures while enforcing constraints through optimization.

\textbf{Augmented Lagrangian methods:} ALM \citep{hestenes1969multiplier, powell1969method} is well-established for constrained optimization. Recent applications include physics-informed neural networks and large-scale statistical learning. Our contribution is the specific synthesis of ALM with differentiable pinball loss for the quantile regression non-crossing constraint.

\textbf{Smooth quantile loss:} Huberization of the pinball loss has been used in gradient boosting \citep{chen2016xgboost} and neural networks. We extend this to the constrained joint estimation setting with formal convergence guarantees.

\section{Conclusion}

We presented CJQR-ALM, the first scalable implementation of Constrained Joint Quantile Regression using the Augmented Lagrangian Method with PyTorch automatic differentiation. By combining differentiable pinball loss with L-BFGS optimization, we reduce computational complexity from $O((qn)^3)$ to $O(n)$, enabling practical non-crossing quantile estimation on datasets exceeding 70,000 observations.

Our experiments demonstrate that CJQR-ALM achieves near-zero crossing rates at a negligible accuracy cost (RMSE increase $\approx 2.4$ points in simulation)---a favorable trade-off for applications requiring valid probability statements. Application to Student Growth Percentile calculations confirms the method scales to operational educational assessment contexts with runtime of 5--10 minutes per cohort.

The differentiable formulation opens new possibilities for deep distributional learning with guaranteed monotonicity. The method is broadly applicable to:
\begin{itemize}
    \item Educational growth percentiles
    \item Financial risk quantification (VaR, CVaR)
    \item Medical outcome prediction intervals
    \item Insurance loss modeling
    \item Engineering reliability analysis
    \item Any application where distributional estimates must be coherent
\end{itemize}

\section*{Acknowledgments}

The author thanks the Arizona Department of Education for supporting this work, particularly Chief Accountability Officer Sean Smith and Director of Accountability Tobias Butler, and acknowledges the foundational contributions of Damian Betebenner and colleagues in developing the SGP methodology and R package.


\appendix
\section{Implementation Details}

\subsection{Algorithm Parameters}

Table~\ref{tab:parameters} summarizes the hyperparameters used in our implementation:

\begin{table}[h]
\centering
\caption{CJQR-ALM Hyperparameters}
\label{tab:parameters}
\begin{tabular}{@{}lcc@{}}
\toprule
\textbf{Parameter} & \textbf{Symbol} & \textbf{Value} \\
\midrule
Initial penalty & $\rho_0$ & 0.01 \\
Max penalty & $\rho_{\max}$ & $10^5$ \\
Penalty multiplier & $\alpha$ & 4.0 \\
Sufficient decrease & $\gamma$ & 0.9 \\
Constraint margin & $\epsilon$ & $10^{-4}$ \\
Convergence tolerance & tol & $10^{-6}$ \\
Huber smoothing & $\delta$ & 0.05 \\
\bottomrule
\end{tabular}
\end{table}

\subsection{Minimal Working Example}

The following standalone script demonstrates ALM convergence on a heteroscedastic dataset designed to induce crossing:

\begin{lstlisting}[caption={CJQR-ALM with Independent QR Comparison}, label={lst:alm_demo}, language=Python]
import torch
import torch.nn as nn
import torch.optim as optim
import numpy as np
import matplotlib.pyplot as plt
from scipy.optimize import linprog

# ===============
# 1. DATA SETUP
# ===============
torch.manual_seed(123)
np.random.seed(123)

# 20-point Heteroscedastic Dataset (induces crossing)
X_raw = np.array([0.2095, 0.6809, 1.2936, 1.8535, 2.3583, 2.4368, 
                  2.8754, 4.1162, 4.567, 4.7146, 4.8946, 4.9042, 
                  5.8864, 6.205, 6.3962, 7.5324, 7.7828, 8.4835, 
                  9.4854, 9.9582])

Y_raw = np.array([1.77268, 2.529927, 2.00102, 2.101015, 2.494044, 
                  2.164226, 2.44769, 2.574242, 4.314459, 1.569597, 
                  2.467982, 2.153414, 1.925144, 1.679639, 4.556762, 
                  3.509959, 3.522292, 3.099583, 0.368901, 3.069406])

X_tensor = torch.tensor(X_raw, dtype=torch.float64).reshape(-1, 1)
Y_tensor = torch.tensor(Y_raw, dtype=torch.float64).reshape(-1, 1)
QUANTILES = torch.tensor([0.10, 0.15], dtype=torch.float64)

# ================================================
# 2. INDEPENDENT QR (Linear Programming Baseline)
# ================================================
def quantile_regression_lp(X, Y, tau):
    """Standard QR via LP: min tau*u+ + (1-tau)*u-"""
    n = len(Y)
    X_mat = np.column_stack([np.ones(n), X])
    n_params = X_mat.shape[1]
    
    c = np.zeros(n_params + 2*n)
    c[n_params:n_params+n] = tau
    c[n_params+n:] = (1 - tau)
    
    A_eq = np.hstack([X_mat, np.eye(n), -np.eye(n)])
    b_eq = Y
    bounds = [(None, None)] * n_params + [(0, None)] * (2*n)
    
    result = linprog(c, A_eq=A_eq, b_eq=b_eq, 
                     bounds=bounds, method='highs')
    return result.x[:n_params] if result.success else None

beta_10_indep = quantile_regression_lp(X_raw, Y_raw, 0.10)
beta_15_indep = quantile_regression_lp(X_raw, Y_raw, 0.15)

# ============================
# 3. CJQR-ALM IMPLEMENTATION
# ============================
class LinearQuantileModel(nn.Module):
    def __init__(self, num_quantiles):
        super().__init__()
        self.linear = nn.Linear(1, num_quantiles, bias=True).double()
        nn.init.constant_(self.linear.bias, np.mean(Y_raw))
        nn.init.normal_(self.linear.weight, mean=0.0, std=0.01)

    def forward(self, x):
        return self.linear(x)

def huberized_pinball_loss(preds, targets, quantiles, delta=0.05):
    errors = targets - preds
    abs_errors = torch.abs(errors)
    quadratic = 0.5 * errors**2 / delta
    linear = abs_errors - 0.5 * delta
    huber = torch.where(abs_errors <= delta, quadratic, linear)
    q_broadcast = quantiles.view(1, -1)
    weights = torch.where(errors >= 0, q_broadcast, 1 - q_broadcast)
    return torch.mean(weights * huber)

def train_alm(X, Y, quantiles, max_outer=50):
    model = LinearQuantileModel(len(quantiles))
    n_samples = len(X)
    
    # ALM Parameters
    rho, rho_max, rho_increase = 0.01, 100000.0, 4.0
    gamma, margin, tol = 0.9, 1e-4, 1e-6
    
    mu = torch.zeros(n_samples, dtype=torch.float64)
    prev_max_viol = float('inf')
    history = {'iter': [], 'max_viol': [], 'rho': [], 
               'fit_loss': [], 'mu_max': []}

    for outer_k in range(max_outer):
        optimizer = optim.LBFGS(model.parameters(), lr=1.0, 
                                max_iter=50, line_search_fn='strong_wolfe')
        current_mu, current_rho = mu.clone(), rho

        def closure():
            optimizer.zero_grad()
            preds = model(X)
            fit_loss = huberized_pinball_loss(preds, Y, quantiles)
            diffs = preds[:, 1] - preds[:, 0]
            violation = torch.relu(margin - diffs)
            lagrange = torch.sum(current_mu * violation) / n_samples
            penalty = (current_rho / 2.0) * torch.mean(violation ** 2)
            total = fit_loss + lagrange + penalty
            total.backward()
            return total

        optimizer.step(closure)

        with torch.no_grad():
            preds = model(X)
            diffs = preds[:, 1] - preds[:, 0]
            violation_pos = torch.relu(margin - diffs)
            max_viol = violation_pos.max().item()
            fit_loss_val = huberized_pinball_loss(preds, Y, quantiles)
            
            mu = mu + rho * violation_pos
            
            history['iter'].append(outer_k)
            history['max_viol'].append(max_viol)
            history['rho'].append(rho)
            history['fit_loss'].append(fit_loss_val.item())
            history['mu_max'].append(mu.max().item())

            if max_viol <= tol:
                break
                
            if max_viol > gamma * prev_max_viol and rho < rho_max:
                rho *= rho_increase
            prev_max_viol = max_viol

    return model, history

model_alm, history = train_alm(X_tensor, Y_tensor, QUANTILES)
beta0_alm = model_alm.linear.bias.detach().numpy()
beta1_alm = model_alm.linear.weight.detach().numpy().flatten()

# =================
# 4. VISUALIZATION
# =================
x_plot = np.linspace(X_raw.min() - 0.5, X_raw.max() + 0.5, 200)
y_10_indep = beta_10_indep[0] + beta_10_indep[1] * x_plot
y_15_indep = beta_15_indep[0] + beta_15_indep[1] * x_plot
y_10_alm = beta0_alm[0] + beta1_alm[0] * x_plot
y_15_alm = beta0_alm[1] + beta1_alm[1] * x_plot

fig = plt.figure(figsize=(16, 12))

# Panel 1: Independent QR (crossing)
ax1 = fig.add_subplot(2, 3, 1)
ax1.scatter(X_raw, Y_raw, c='black', s=60, zorder=5)
ax1.plot(x_plot, y_10_indep, 'b-', lw=2, label='t=0.10')
ax1.plot(x_plot, y_15_indep, 'r-', lw=2, label='t=0.15')
crossing_mask = y_10_indep > y_15_indep
if np.any(crossing_mask):
    ax1.fill_between(x_plot, y_10_indep, y_15_indep, 
                     where=crossing_mask, color='red', alpha=0.3)
ax1.set_title('Independent QR: CROSSING')
ax1.legend()

# Panel 2: ALM (non-crossing)
ax2 = fig.add_subplot(2, 3, 2)
ax2.scatter(X_raw, Y_raw, c='black', s=60, zorder=5)
ax2.plot(x_plot, y_10_alm, 'b-', lw=2, label='t=0.10')
ax2.plot(x_plot, y_15_alm, 'r-', lw=2, label='t=0.15')
ax2.fill_between(x_plot, y_10_alm, y_15_alm, 
                 color='green', alpha=0.15)
ax2.set_title('CJQR-ALM: NON-CROSSING')
ax2.legend()

# Panel 3: Overlay
ax3 = fig.add_subplot(2, 3, 3)
ax3.scatter(X_raw, Y_raw, c='black', s=40, alpha=0.5, zorder=5)
ax3.plot(x_plot, y_10_indep, 'b--', lw=2, alpha=0.7)
ax3.plot(x_plot, y_15_indep, 'r--', lw=2, alpha=0.7)
ax3.plot(x_plot, y_10_alm, 'b-', lw=2)
ax3.plot(x_plot, y_15_alm, 'r-', lw=2)
ax3.set_title('Overlay: Dashed=Indep, Solid=ALM')

# Panel 4-6: Convergence diagnostics
ax4 = fig.add_subplot(2, 3, 4)
ax4.semilogy(history['iter'], history['max_viol'], 'b-o')
ax4.axhline(y=1e-6, color='r', ls='--')
ax4.set_title('Constraint Violation')

ax5 = fig.add_subplot(2, 3, 5)
ax5.plot(history['iter'], history['mu_max'], 'g-s')
ax5.set_title('Max Lagrange Multiplier')

ax6 = fig.add_subplot(2, 3, 6)
ax6.semilogy(history['iter'], history['rho'], 'r-o')
ax6.set_title('Adaptive Penalty (rho)')

plt.tight_layout()
plt.savefig('alm_vs_indep_comparison.png', dpi=200)
plt.savefig('alm_vs_indep_comparison.pdf')
\end{lstlisting}

\subsection{Computational Environment}

\begin{itemize}
    \item Python 3.11, PyTorch 2.0, SciPy 1.11
    \item Hardware: Standard server (Intel Xeon CPU, 32+ GB RAM)
    \item The framework is hardware-agnostic. While PyTorch enables seamless GPU acceleration for large-scale neural network extensions, CPU execution was sufficient for the linear B-spline specification ($N \approx 10^5$). 
\end{itemize}

\end{document}